\newif\ifjournal
\journalfalse % Initial and revised submission

\ifjournal
    \documentclass[letterpaper, 10 pt, journal, twoside]{IEEEtran} % Use this command for final RAL version

    \newenvironment{keywords}{\begin{IEEEkeywords}}{\end{IEEEkeywords}}    

\else
    \documentclass[letterpaper, 10 pt, conference]{ieeeconf}
    \IEEEoverridecommandlockouts % This command is only needed if you want to use the \thanks command
    \newcommand{\IEEEPARstart}[2]{{#1#2}}

    \makeatletter %https://tex.stackexchange.com/questions/160109/citations-not-linking-to-bibliography
    \let\NAT@parse\undefined
    \makeatother
\fi

\usepackage{graphicx}
\usepackage{hyperref} % hyperlinks for urls and references
\usepackage{spverbatim} % verbatim with line breaks
\usepackage{amsmath}
\usepackage{amssymb}
\usepackage[caption=false, font=footnotesize, subrefformat=parens,labelformat=parens]{subfig}
\usepackage{booktabs}
\usepackage{flushend}

\graphicspath{{figures}}

\DeclareMathOperator*{\argmin}{arg\,min}
\def\rot{\textit{\textsf R}}
\def\norm#1{\left\|{#1}\right\|}
\def\tr{^T}
\DeclareMathOperator{\sign}{sign}

\usepackage{xcolor}

\title{
\ifjournal\else
    \LARGE\bf
\fi
Redundancy Parameterization of the ABB YuMi Robot Arm}

\ifjournal
\fi
\author{Alexander~J.~Elias,~\IEEEmembership{Graduate Student Member,~IEEE}, and John~T.~Wen,~\IEEEmembership{Life~Fellow,~IEEE}% <- Prevent space
\thanks{\textit{(Corresponding author: Alexander J. Elias)}}
\ifjournal
\thanks{Manuscript received: Month Day, Year; Revised: Month Day, Year; Accepted: Month Day, Year.}%
\thanks{This paper was recommended for publication by Editor FirstName A. EditorName upon evaluation of the Associate Editor and Reviewers’ comments.}%
\fi
\thanks{The authors are with the Department of Electrical, Computer, and Systems Engineering, Rensselaer Polytechnic Institute, Troy, NY 12180 USA (e\mbox{-}mail:~eliasa3@rpi.edu; wenj@rpi.edu).}%
}

\begin{document}

\maketitle
\begin{abstract} % RA-L: no more than 200 words, 1800 characters
The ABB YuMi is a 7-DOF collaborative robot arm with a complex, redundant kinematic structure. Path planning for the YuMi is challenging, especially with joint limits considered. The redundant degree of freedom is parameterized by the Shoulder-Elbow-Wrist (SEW) angle, called the arm angle by ABB, but the exact definition must be known for path planning outside the RobotStudio simulator.
We provide the first complete and validated definition of the SEW angle used for the YuMi. It follows the conventional SEW angle formulation with the shoulder-elbow direction chosen to be the direction of the fourth joint axis.
Our definition also specifies the shoulder location, making it compatible with any choice of reference vector.
A previous attempt to define the SEW angle exists in the literature, but it is incomplete and deviates from the behavior observed in RobotStudio.
Because our formulation fits within the general SEW angle framework, we also obtain the expression for the SEW angle Jacobian and complete numerical conditions for all algorithmic singularities.
Finally, we demonstrate using IK-Geo, our inverse kinematics (IK) solver based on subproblem decomposition, to find all IK solutions using 2D search.
Code examples are available in a publicly accessible repository.
\end{abstract}
\section{Introduction} 
\IEEEPARstart{D}{espite} being launched over 10 years ago in 2015, path planning for the ABB YuMi 7-DOF collaborative robot remains a challenge. Path planning is difficult because of the combination of a complex kinematic structure with no consecutive intersecting or parallel axes, a redundant degree of freedom, and restrictive joint limits. The YuMi is not the only arm with this kinematic structure: The Robotics Research Corporation (RRC) arms analyzed in \cite{kreutz1992kinematic} also had this structure and were some of the first 7-DOF arms to be analyzed and considered for use in space. These arms were never flown in space, perhaps because of difficulties with kinematics, and instead arms used in space tend to have three consecutive intersecting or parallel joint axes.

Moreover, the kinematic model of the YuMi has not been fully described before. There is no publicly available description of the forward kinematics including the redundancy parameterization which matches the ABB controller. There is also no inverse kinematics (IK) method which has been demonstrated to find all IK solutions matching the ABB controller. 

There are two versions of the YuMi: the IRB 14000 (dual-arm) and IRB 14050 (single-arm). We focus on the single-arm version. Although our analysis will be performed in ABB's simulator RobotStudio, the controller matches the behavior of the real robot.

Since the YuMi is a 7-DOF arm, there is a redundant degree of freedom which must be parameterized to fully specify the arm configuration for a given end effector pose. For most end effector poses, the redundant degree of freedom looks like the rotation of the robot elbow about the line from the shoulder to the wrist, so a good redundancy parameterization is able to assign an angle to each position of the elbow along this self-motion manifold. In the ABB controller, the redundant degree of freedom of the YuMi is parameterized by the Shoulder-Elbow-Wrist (SEW) angle, called the arm angle by ABB, and sometimes called the swivel angle.

An understanding of the SEW angle is important for several reasons, including not only intuitive and accurate programming and teleoperation but also offline path planning.
When doing path planning it is important to know the singularities, including the algorithmic singularities arising from the definition of the SEW angle. For a given task-space path, there may or may not be a corresponding joint-space path. If there are multiple joint-space paths, a path planner should determine the best joint-space path. The feasibility of a given task-space path depends on the presence of singularities. For redundant manipulators with a redundancy parameterization such as the ABB YuMi, there are two broad classes of singularities: kinematic singularities and algorithmic singularities.

Crossing a kinematic singularity (or, more precisely, passing through a singular pose in the singular direction) is not possible because of the physical limitations of the arm. When the rank of the \(6\times7\) kinematic Jacobian decreases to 5, there is some end effector velocity which is not possible. This also means there are 2 degrees of freedom for self-motion, so one scalar parameterization alone cannot parameterize the redundancy. Moving in the singular direction might involve moving certain joints infinitely fast to reconfigure the robot, or it may be impossible because the robot would have to move outside the boundary of its workspace. 

Algorithmic singularities, on the other hand, occur not because of the physical limitations of the robot but because of the algorithm used to parameterize the redundancy. At an algorithmic singularity, there may be some self-motion which is not captured by the parameterization (augmentation singularity). Or, at an algorithmic singularity the Jacobian of the parameterization may be undefined, which typically occurs due to a division by zero. In rare cases, the Jacobian of the parameterization is zero.

Even if a path does not cross a singularity, proximity to a singularity can significantly affect behavior. Near kinematic singularities, the \(6\times7\) kinematic Jacobian becomes ill-conditioned which means that certain task velocities result in large joint velocities. Near algorithmic singularities, the sensitivity between the redundancy parameter velocity and the self-motion becomes very small or very large.
A small change in end effector pose while keeping the redundancy parameter constant may result in large self-motion.
A small change in the redundancy parameter may result in large self-motion, or a small self-motion may result in a large change in the redundancy parameter.

In order to properly perform path planning for the YuMi, we need a full understanding of both the kinematic singularities as well as the SEW angle definition which results in the algorithmic singularities. However, the ABB manuals do not provide a complete or fully correct explanation for the kinematic singularities or how the SEW angle is calculated. Only some of the kinematic singularities are discussed, and there is no explanation provided for how to calculate the SEW angle.

The kinematic and algorithmic singularities of the YuMi were recently described thoroughly for the first time in \cite{asgari2025singularities}. The authors used Grassman line geometry to systematically identify and visualize all kinematic singularities. (A similar technique was used in their earlier paper to identify all singularities of the 6-DOF Kinova arm \cite{kinovaSingularity}.) The analysis reveals there are many complex cases for kinematic singularities of the YuMi which are not discussed in the ABB manuals. Their work serves as a model example of systematically identifying kinematic singularities.

The authors also proposed for the first time the formula for the SEW angle used by ABB, along with a proposed list of algorithmic singularities which occur due to this definition. In our testing, this SEW angle definition appeared correct based on preliminary comparisons with RobotStudio. However, after further testing, especially near predicted singularities, it was discovered that the SEW angle definition was incorrect. This discrepancy was difficult to spot because the SEW angle definition gave correct results down to the second decimal place for much of the workspace. However, there are critical differences between the two definitions. There are not only small numerical deviations but also differences in the singularity structure which create meaningful differences for path planning. The algorithmic singularities were also not fully explored.

The goal of this paper is to extend the excellent work of \cite{asgari2025singularities} to improve the understanding of the YuMi's SEW angle and its singularities.
We present for the first time the correct and complete SEW angle definition of the ABB YuMi. This definition fits within the general SEW angle framework presented in \cite{elias2024_7dof}: It is the conventional SEW angle with the shoulder-elbow direction defined by the direction of joint axis 4 plus a quarter-turn offset. This means we can calculate the forward kinematics, Jacobian, and inverse kinematics, finding all solutions with a 2D search.

In our formulation, we specify the shoulder location so that the definition applies to any choice of reference vector. We find the shoulder is in a different location than specified by ABB. The definition in \cite{asgari2025singularities} does not explicitly specify the shoulder location and therefore only applies to the reference vector pointing along the first joint axis.

The correctness of our SEW angle definition is confirmed with several test cases comparing to RobotStudio. While many test cases do not fail for the definition from~\cite{asgari2025singularities}, some hand-picked cases show a difference.

We discuss the algorithmic singularity structure of the SEW angle used by ABB, and we compare to the SEW angle definition proposed in \cite{asgari2025singularities}. We briefly discuss the possibility of modifying the SEW angle definition to use the Stereographic SEW angle~\cite{elias2024_7dof}, which reduces the algorithmic singularity associated with the wrist being along the line spanned by the reference direction to only a half-line, which can be placed in the robot base and out of reach.

The remainder of this paper is organized as follows.
We describe the kinematic parameters of the ABB YuMi, including the shoulder and wrist positions, using coordinate-free and product of exponentials notation in Section~\ref{sec:kin_params}.
We explain the previous discussions of SEW angle by ABB and by~\cite{asgari2025singularities} in Section~\ref{sec:prior_work}.
Then, we provide the correct SEW angle definition in Section~\ref{sec:SEW_ABB_def} and discuss the resulting singularities in Section~\ref{sec:singularities}.
IK using this SEW angle definition is explained in Section~\ref{sec:IK}.
We confirm the correctness of our SEW angle definition and IK method by comparing to RobotStudio in Section~\ref{sec:testing},
and we conclude in Section~\ref{sec:conclusion}.

Code examples are available in a publicly accessible repository.\footnote{\url{https://github.com/rpiRobotics/yumi-ik}}

\section{YuMi Kinematic Parameters}\label{sec:kin_params}
\begin{figure}[t]
    \centering
    \includegraphics[scale = 0.5, clip]{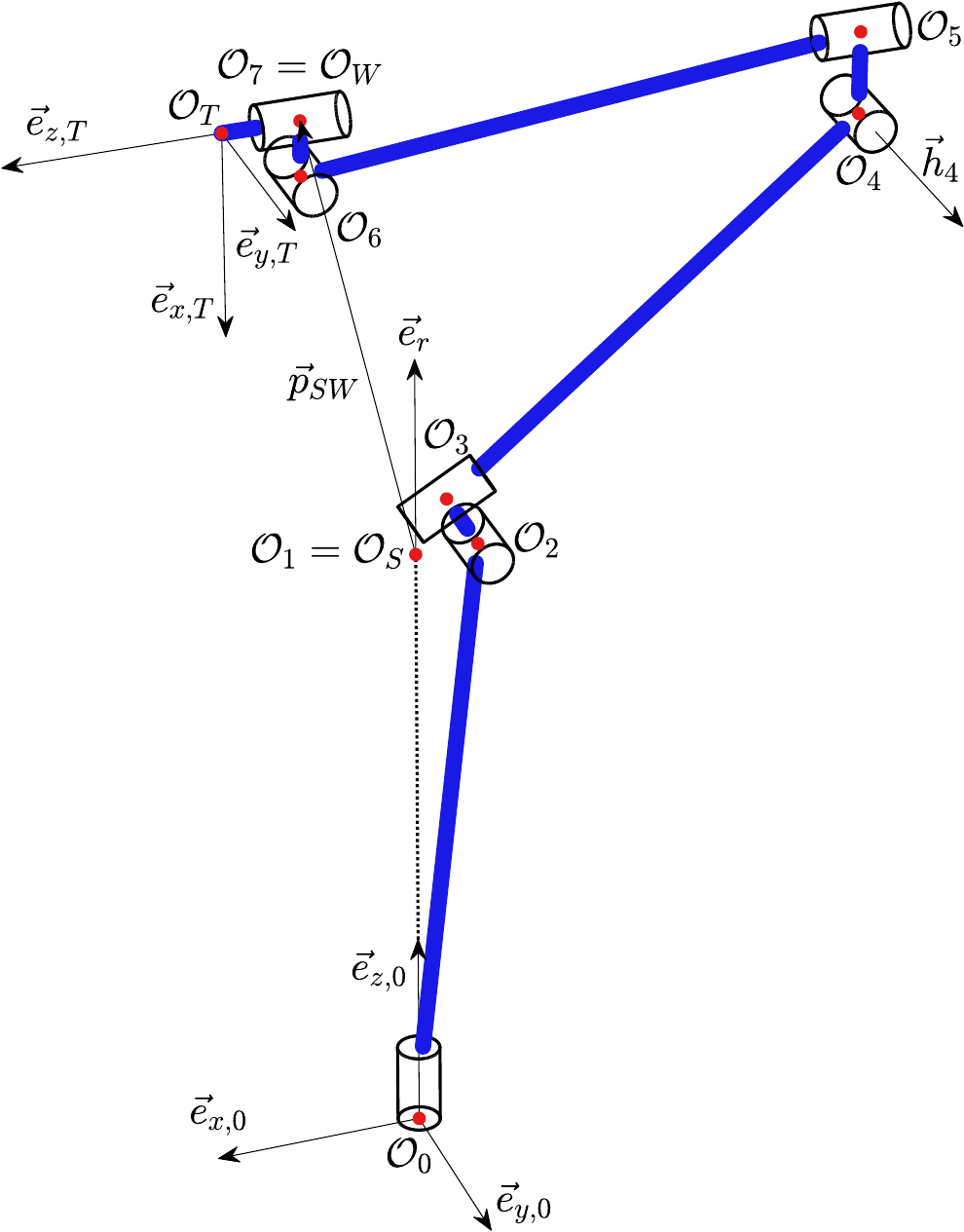}
    \caption{Kinematics of YuMi including locations for shoulder \(\mathcal O_S\) and wrist \(\mathcal O_W\). The SEW angle is calculated using the direction of joint axis 4 \(\vec h_4\), so no elbow point is defined. In this case, the reference vector \(\vec e_r\) is chosen to be equal to \(\vec e_{z,0} = \vec h_1\). The joint angle vector is \(q = [0\ {-\pi}/3\ 0\ \pi/3\ 0\ 0\ 0]\). The unit vectors \(\vec e_i\) are shown with length 100 mm.}
    \label{fig:yumi_kinematics}
\end{figure}

We will use coordinate-free notation and the product of exponentials approach to describe the robot kinematics, as was done in \cite{elias2024_7dof}. Denote a vector in space as \(\vec p\) and a point in space as \(\mathcal O\). Denote an orthonormal frame as \(\mathcal E = [ \vec e_x \  \vec e_y \ \vec e_z]\). Commonly, we consider a frame along with a frame origin~\((\mathcal E, \mathcal O)\). Denote the rotation operator about \(\vec h\) over angle \(\theta\) as  \(\mathcal R(\vec h,\theta)\). To express a coordinate-free object in a coordinate system \(\mathcal E\), we use the adjoint operator 
\begin{equation}
    \mathcal E^*=\begin{bmatrix} \vec e_x \cdot \\ \vec e_y \cdot \\ \vec e_z \cdot \end{bmatrix}.
\end{equation}
Vector~\(\vec p\) represented in frame~\(\cal E\) is denoted \(p = \mathcal E^* \vec p\). 
Frame~\(\mathcal E_b\) represented in frame~\(\mathcal E_a\) is the \(SO(3)\) matrix~\(R_{ab}=\mathcal E_a^* \mathcal E_b\). Rotation~\(\mathcal R(\vec h,\theta)\) in frame~\(\mathcal E_a\) is the \(SO(3)\) matrix~\(\rot(h_a,\theta)=e^{h_a^\times \theta}\), where \((\cdot)^\times\) is the~\(3\times 3\) skew-symmetric matrix representation of the cross product:
\begin{equation}
    h^\times := \begin{bmatrix}  0 & -h_z & h_y \\ h_z & 0 & -h_x \\ -h_y & h_x & 0 \end{bmatrix}.
\end{equation}
For a unit vector \(h\), \(-h^{\times^2} = -h^\times h^\times = I - h h\tr\) is the projector onto the orthogonal complement of \(h\).  In coordinate free notation, we have the similar equation \(-(\vec h \times)^2 = \mathcal I - \vec h \vec h \cdot\), where \(\mathcal I\) is the identity operator. We denote \(e_x, e_y, e_z\) as the unit vectors, e.g., \(e_x = \mathcal E^\star \vec e_x = [1\ 0 \ 0]\tr\).

To describe robot kinematics using the product of exponentials approach, we assign frames and origins to each link along the kinematic chain and describe their relative poses.
Denote the robot base frame and origin as \((\mathcal E_0, \mathcal O_0)\) and the end effector / tool frame and origin as \((\mathcal E_T, \mathcal O_T)\). Pick any point along joint axis \(i\) as the link frame origin \(\mathcal O_i\), where \(i \in \{1, 2, \dots, 7\}\) for a 7-DOF robot. Pick an arbitrary robot configuration to use as the zero configuration where the robot joint angle vector \(q = [q_1\ q_2\ \dots \ q_7]\tr=0\). At this zero configuration, all link frames \(\mathcal E_i\) are equal to the base frame~\(\mathcal E_0\).

For a revolute joint \(i\) which rotates link \(i\), denote the joint rotation axis\footnote{Unlike other conventions to describe robot kinematics, the joint rotation axis does not have to point along the \(z\) axis of the link frame.} as \(\vec h_i\), which is constant in the link frame \(\mathcal E_i\) (and \(\mathcal E_{i-1}\)). The link vector \(\vec p_{i, i+1}\) pointing from \(\mathcal O_i\) to \(\mathcal O_{i+1}\) is also constant in \(\mathcal E_i\). For a joint angle \(q_i\), the orientation of frame~\(i\) is given by \(\mathcal E_i=\mathcal R(\vec h_i, q_i)\mathcal E_{i-1}\). 

The kinematic parameters for a robot are given by \(\left(\{p_{i-1,i}\}_{i=1}^7,\ p_{7T},\ \{h_i\}_{i=0}^7,\ R_{7T}\right)\). The link vector \(p_{i-1,i}=\mathcal E_{i-1}^* \vec p_{i-1, i}\) can be found by reading off the offset from \(\mathcal O_{i-1}\) to \(\mathcal O_{i}\) in the base frame when the robot is in the zero configuration. The tool offset vector \(p_{7T} = \mathcal E_7^* p_{7T}\) and the joint axis vector \(h_i = \mathcal E_i^* \vec h_i\) can also be read off in the base frame when the robot is in the zero configuration. The constant tool frame rotation with respect to frame 7 is \(R_{7T}\), which is equal to the tool frame rotation with respect to the base frame when the robot is in the zero pose.

The forward kinematics is the mapping from the base frame to the task frame and is given by 
\begin{subequations}\label{eq:fwdkin}
\begin{align}
R_{0T} ={}& R_{01}R_{12}R_{23}R_{34}R_{45}R_{56}R_{67}R_{7T},\\
\begin{split}
p_{0T} ={}& p_{01}+R_{01}p_{12}+R_{02}p_{23}+R_{03}p_{34} 
\\&{}+ R_{04}p_{45} + R_{05} p_{56} + R_{06} p_{67} + R_{07} p_{7T},
\end{split}
\end{align}
\end{subequations}
where \(R_{i-1, i} = \rot(h_i, q_i)\) and \(R_{i,j} = R_{i,i+1}\cdots R_{j-1,j}\) for~\(i<j\). We also denote \(R_{j,i} = R_{i,j}\tr\) for~\(i<j\).

The kinematic parameters of the ABB YuMi, where \(e_x, e_y, e_z\) are the unit vectors (expressed in mm), are
\begin{equation}\label{eq:kinematic_params}
\begin{gathered}
    p_{01} = 306 e_z,\
    p_{12} = -30 e_x,\
    p_{23} = 30 e_x,\\
    p_{34} = 40.5 e_x + 251.5 e_z,\
    p_{45} = 40.5 e_z,\\
    p_{56} = 265 e_x + 27 e_z,\
    p_{67} = 27 e_z,\
    p_{7T} = 36 e_x,\\
    R_{7T} = \rot(e_y, \pi/2),\\
    h_5 = h_7 = e_x,\
    h_2 = h_4 = h_6 = e_y,\
    h_1 = h_3 = e_z.
\end{gathered}
\end{equation}
The kinematics are illustrated in Fig.~\ref{fig:yumi_kinematics}. There are no consecutive intersecting or parallel joint axes, although joints \((1, 3)\), \((3, 5)\), and \((5, 7)\) have nonconsecutive intersecting axes.

Joint limits, in degrees, are given by
\(
    q_{min} = [{-168.5}\ {-143.5}\  {-168.5}\ {-123.5}\ {-290}\ {-88}\ {-229}]\) and
\(
    q_{max} = [168.5\ 43.5\   168.5\ 80\    290\ 138\  229]\tr.
\)
There are poses within the joint limits for which the robot still has self-intersections. Joints 5 and 7 have more than one full revolution of motion possible, so care must be taken during path planning to consider IK solutions with \(2\pi\) offsets for each joint.

The joint numbering is different in RobotStudio than in the product of exponentials model because \(q_3\) is moved to the end. The joint angle vector in RobotStudio is~\(\begin{bmatrix}
q_1 \ q_2 \ q_4 \ q_5 \ q_6 \ q_7 \ q_3
\end{bmatrix}\tr\). 

In order to define the SEW angle, we must also specify the shoulder and wrist points. For the ABB YuMi, we have shoulder position \(\mathcal O_S = \mathcal O_1\) and wrist position \(\mathcal O_W = \mathcal O_7\). (While the link origins can be placed arbitrarily along \(\vec h_1\) and \(\vec h_7\), the shoulder and wrist positions cannot.)

This is the first time the YuMi shoulder position has been described. The shoulder position does not precisely match the description given by ABB in the manuals. ABB says the shoulder passes through the origin of joint axis 2~\cite{ABB_tech_ref_manual, ABB_YUMI}. However, the true shoulder is the point on axis 1 closest to joint axis 2. Only one choice of SEW reference vector was used in \cite{asgari2025singularities}, so the shoulder position was arbitrary along a line.

As noted by \cite{asgari2025singularities}, the location of the wrist, called the Wrist Center Point (WCP) by ABB, is not explicitly discussed in the manuals, but it is the point on joint axis 7 closest to joint axis 6.

The SEW angle is defined using the fourth axis direction \(\vec h_4\), which represented in the base frame is given by \(R_{03}h_4\).
This is a special case of how to define the SEW angle, as a more common choice is to pick some point on the kinematic chain as the elbow point \(\mathcal O_E\) and use the shoulder-elbow vector \(p_{SE}\). However, as demonstrated in~\cite{elias2024_7dof}, using \(\vec h_4\) is a valid and sometimes advantageous alternative, equivalent to placing \(\mathcal O_E\) infinitely far away in the  \(\vec h_4\) direction.

Define \(\mathcal E_C\) as the frame used to measure the SEW angle. The \(z\) vector in this frame \(\vec e_{z,C}\) points in the same direction as the shoulder-wrist vector \(\vec p_{SW}\). The remaining degree of freedom for the frame orientation, corresponding to the choice of \(\vec e_{x,C}\) and \(\vec e_{y,C}\), depends on the choice of the SEW angle reference direction \(\vec e_r\) and will be described fully in the next sections. The ABB controller allows users to choose the reference vector \(\vec{e}_r\) as one of the following: the first joint axis \(\vec{h}_1\) (which is also \(\vec{e}_{z,0}\)), the world \(z\)-axis (\(\vec{e}_{z,0}\) in the default base pose), or the world \(y\)-axis (\(\vec{e}_{y,0}\) in the default pose).

The coordinate representations \(e_{x,C}\),  \(e_{y,C}\), \(e_{z,C}\), \(p_{SW}\), and \(e_r\) are all expressed in the base frame~\(\mathcal E_0\).

\section{Prior Work}\label{sec:prior_work}
\subsection{ABB Manuals}
In their manuals~\cite{ABB_tech_ref_manual, ABB_YUMI, RAPID_overview}, ABB does not provide a complete explanation of how the SEW angle is calculated. They state that the SEW angle ``cannot be measured or calculated by the user since the
underlying mathematical calculations are far more complex"~\cite{ABB_YUMI}. However, as we will see in Section~\ref{sec:SEW_ABB_def}, the SEW angle calculation can be visualized and has a compact closed-form expression.

The manuals also discuss singularities. They explain the YuMi has all the 6-DOF arm singularities: The wrist center intersecting axis 1, the wrist singularity at \(q_6=0\) where joints 5 and 7 become collinear, and singularity when the arm outstretched. The manuals also explain there are two additional singularities for the YuMi: The \(q_2=0\) shoulder singularity where joints 1 and 3 become collinear and the singularity where the wrist is on the line passing through the shoulder in the SEW angle reference direction. However, this explanation is incomplete and, in some cases, incorrect. The YuMi does not encounter a singularity when the wrist passes through axis 1, provided the reference direction is chosen appropriately. As \cite{asgari2025singularities} showed, there are many more cases of kinematic singularities. As we will show in Section~\ref{sec:singularities}, there are other cases of algorithmic singularities. The manuals do not distinguish between kinematic singularities, which are unavoidable, and algorithmic singularities, which can be avoided by bypassing the SEW angle parameterization entirely, such as by using joint-level control.

While \cite{asgari2025singularities} states that the SEW angle cannot be automatically selected by the ABB controller, the controller can in fact adjust it automatically to avoid singularities.
This behavior is governed by the ``Limit avoidance distance" controller parameter. According to the manuals, the controller avoids the \(q_2=0\) and \(q_6=0\) singularities as well as joint limits. It is unclear whether other singularities are avoided as well.

\subsection{Asgari, Bonev, and Gosselin}
In response to the limited discussion by ABB, the authors of \cite{asgari2025singularities} were the first to begin fully describing the YuMi singularities and SEW angle.
In this paper they said the SEW definition was the same as the conventional SEW angle described in~\cite{kreutz1992kinematic}, although with angle measurement made using the joint 4 axis direction. We agree with this statement. However, the equations and geometric descriptions provided in~\cite{asgari2025singularities} do not exactly match the conventional SEW angle definition. While it appears the definition matches RobotStudio for many test cases within the joint limits, their definition deviates from that used by ABB and results in a completely different singularity structure. Furthermore, the SEW angle definition in~\cite{asgari2025singularities} only works when the reference vector points along the first joint axis as the shoulder position is not explicitly defined.

The definition from \cite{asgari2025singularities}, which we find does not match RobotStudio, is expressed using our notation as
\begin{equation}\label{eq:SEW_ABB}
    \psi^{sign} = \mbox{ATAN2}(
    \sigma \norm{(-e_y)^\times R_{03}h_4},
    (-e_y)\tr R_{03}h_4
    ),
\end{equation}
where the sign term is
\begin{equation}
    \sigma = \sign(e_r\tr {R_{03}h_4})
\end{equation}
and \(\vec e_{y,C}\) is the normalized version of \(\vec p_{SW} \times \vec e_r\), that is,
\begin{equation}
    e_{y,C} = \frac
    {
        p_{SW}^\times e_r
    }{\norm{p_{SW}^\times e_r}}
\end{equation}
This is interpreted as
\begin{align}
    \psi^{sign} &= \sigma \angle(-e_{y,C}, R_{03}h_4)\\
    &=\sign(\vec {e_r} \cdot \vec h_4) \angle(-\vec e_{y,C}, \vec h_4),
\end{align}
where \(\angle(\cdot, \cdot)\) is the angle between two vectors.

The equation provided in~\cite{asgari2025singularities} did not show where the shoulder was. When the reference direction points along the joint 1 axis, the shoulder can be placed anywhere along the axis and the calculation for \(e_{y,C}\) and therefore \(\psi^{sign}\) remains the same.

In Section~\ref{sec:testing} we will prove that this SEW angle definition is incorrect by showing discrepancies between this definition and the SEW angle displayed in RobotStudio. This incorrect definition does not fall under the general SEW angle definition from~\cite{elias2024_7dof}.

Beyond the mismatch between \(\psi^{sign}\) and the SEW angle used in RobotStudio, there is also a significant difference in singularity structure. The sign term~\(\sigma\) introduces a singularity whenever \(\vec {e_r} \cdot \vec h_4=0\). Performing path planning using the incorrect SEW angle definition would inadvertently lead to many failed paths, as this singularity condition is common and cannot be avoided with small deviations in the desired task-space path.

\section{SEW Angle for ABB YuMi}\label{sec:SEW_ABB_def}
\begin{figure}[t]
    \centering
    \includegraphics[scale=0.5, clip]{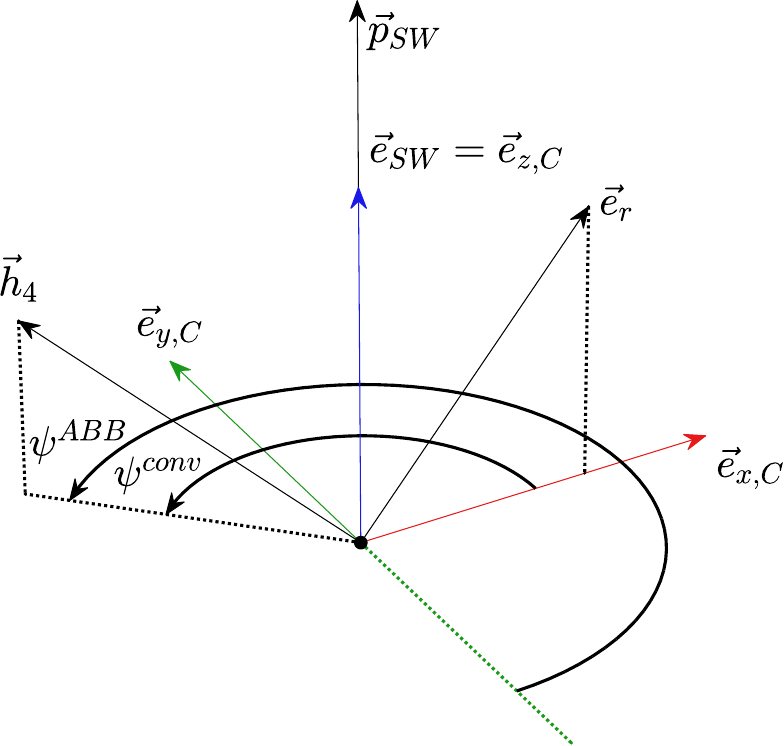}
    \caption{SEW angle calculation. The 4th joint axis direction \(\vec h_4\) and the reference direction \(\vec e_r\) are projected onto the plane normal to the shoulder-wrist direction \(\vec p_{SW}\). The conventional SEW angle is the angle between these two projected vectors. The SEW angle used by ABB is offset by a quarter turn.}
    \label{fig:SEW_calculation}
\end{figure}

The SEW angle used by ABB \(\psi^{ABB}\) is a constant \(\pi/2\) offset from the conventional SEW angle definition \(\psi^{conv}\) where the shoulder-elbow direction is chosen to be \(\vec h_4\). Details of the definition are shown below, and a visualization is shown in Fig.~\ref{fig:SEW_calculation}.

The SEW angle used by the YuMi falls within the general SEW angle definition from \cite{elias2024_7dof}, which is
\begin{align}
    \psi^{gen} &= \argmin_\theta \norm{ \mathcal R(\vec e_{SW}, \theta) \vec e_{x,C} - \vec p_{SE} },
    \\&= \argmin_\theta \norm{ \rot(e_{SW}, \theta) e_{x,C} - p_{SE} },
\end{align}
where \(\vec e_{x,C} = \vec f_x(\vec p_{SW})\) (or \(e_{x,C} = f_x(p_{SW})\) represented in the base frame) is some unit vector orthogonal to \(\vec p_{SW}\) and defines the orientation of \(\mathcal E_C\). The general SEW angle is the angle of the shoulder-elbow direction in frame \(\mathcal E_C\), which is to say it is the angle from \(\vec e_x\) to the projection of \(\vec p_{SE}\) onto the plane normal to \(\vec e_{SW}\) (spanned by \(\vec e_x\) and \(\vec e_y\)) about the rotation axis \(e_{SW}\).

The conventional definition SEW angle definition is
\begin{align}
    \psi^{conv} &= \argmin_\theta \norm{ \mathcal R(\vec e_{SW}, \theta) \vec e_r - \vec p_{SE} }
    \\ &= \argmin_\theta \norm{ \rot(e_{SW}, \theta) e_r - p_{SE} },
\end{align}
where we have chosen the function \(\vec f_x(\cdot)\) so that \(\vec e_{x,C}\) is the normalized projection of \(\vec e_r\) on the plane normal to \(\vec e_{SW}\). For the YuMi we pick \(\vec p_{SE} = \vec h_4\), so \(p_{SE} = R_{03}h_4\).

The SEW angle definition used by the ABB controller is a quarter turn larger than the conventional SEW angle definition:
\begin{equation}
    \psi^{ABB} = \psi^{conv} + \frac{\pi}{2},
\end{equation}
where we wrap \(\psi^{ABB}\) to the range \([-\pi, \pi]\). We equivalently have

\begin{equation}
    \psi^{ABB} = \argmin_\theta \norm{ \rot(e_{SW}, \theta) (-e_y) - R_{03}h_4 },\text{ or}
\end{equation}
\begin{equation}
    \rot(e_{SW}, \psi^{ABB}) (-e_y) = -e_{SW}^{\times^2}R_{03}h_4.
\end{equation}
The extra quarter turn in the ABB definition means we are measuring the angle of \(\vec h_4\) with respect to \(-\vec e_y\) rather than with respect to \(\vec e_x\). Equivalently, this offset can be interpreted as measuring a rotated version of \(\vec h_4\) with respect to \(\vec e_x\), as usually \(\vec e_{SW} \times \vec h_4\) approximately points from the shoulder to the origin of joint 4 or 5.

The angles can be found using Subproblem~1 from \cite{elias2025_6dof}:
\begin{align}
    \psi^{conv} &= \mbox{ATAN2}\left(
    (e_{SW} ^\times e_r)\tr R_{03}h_4,
    -(e_{SW} ^{\times^2} e_r)\tr R_{03}h_4
    \right)\\
    \psi^{ABB} &=  \mbox{ATAN2}\left(-(e_{SW} ^{\times^2} e_r)\tr R_{03}h_4,  -(e_{SW} ^\times e_r)\tr R_{03}h_4\right)\\
    &= \mbox{ATAN2}\left(-e_r\tr e_{SW} ^{\times^2}  R_{03}h_4,  e_r\tr e_{SW}^\times R_{03}h_4\right)
\end{align}

Since the ABB definition \(\psi^{ABB}\) is a constant angle away from the conventional SEW angle \(\psi^{conv}\), the expression for the Jacobian and all the singularity analysis from~\cite{elias2024_7dof} apply.

There are two main differences between the definition provided in this work \(\psi^{ABB}\) and the previous definition \(\psi^{sign}\) which we find does not match RobotStudio. First, \(\psi^{ABB}\) finds the angle between \(-\vec e_{y,C}\) and the projected version of \(\vec h_4\), while \(\psi^{sign}\) does not project \(\vec h_4\) onto the plane normal to \(p_{SW}\). Second, \(\psi^{ABB}\) finds the sign of the angle by considering both the \(x\) and \(y\) axes of the \(C\) frame, while \(\psi^{sign}\) finds the sign of the angle based on if \(\vec h_4\) points towards or away from \(\vec p_{SW}\).

\section{Singularity Analysis}\label{sec:singularities}

With the complete forward kinematics known, including the SEW angle formulation, we can now systematically identify all singularity cases as categorized in~\cite{elias2024_7dof}. This analysis reveals augmentation singularities that have not been previously discussed.

Case~1 is kinematic singularities. This singularity, where the \(6\times 7 \) kinematic Jacobian loses rank, was thoroughly analyzed in~\cite{asgari2025singularities}. The remaining cases below are all algorithmic singularities.

Case~2 is the augmentation singularity, where the kinematic and SEW Jacobians are both defined and full-rank but the augmented Jacobian loses rank, which occurs when self-motion does not change the SEW angle. The ABB manuals and \cite{asgari2025singularities} both discuss the shoulder singularity at \(q_2=0\) and the wrist singularity at \(q_6=0\). In either case, the self motion becomes the counterrotation of joints (1, 3) or (5, 7), which does not change \(\vec h_4\). (\cite{asgari2025singularities} makes the point that \(q_4=0\) is not an algorithmic singularity, as although joints 3 and 5 counterrotate, so this motion is still captured by the SEW angle.)

There are cases of augmentation singularities for the ABB SEW angle beyond the cases of \(q_2=0\) or \(q_6=0\). Graphically, these singular poses can be seen as local extrema (zero-slope points) on graph of \(\psi^{ABB}\) over any other parameterization of self-motion, such as \(q_1\). One example of an augmentation singularity occurs at
\(q = [
  0\ 
  {-31.12}\ 
   61.30\ 
  {-65.33}\ 
 {-132.67}\ 
 {-20.55}\ 
  0]\)~deg.

Case~3 is the SEW angle singularity, where the SEW angle Jacobian is undefined. (Nearby this singularity, the Jacobian norm approaches infinity.) Subcase~3A is the collinear singularity where \(\vec p_{SW}\) is collinear with \(\vec h_4\). This singularity does not appear to occur for the YuMi. Subcase~3B is the coordinate singularity where \(\vec e_r\) is collinear with \(\vec p_{SW}\). This case is discussed by ABB and \cite{asgari2025singularities}.

\begin{figure}[t]
    \centering
    \includegraphics[width=\linewidth]{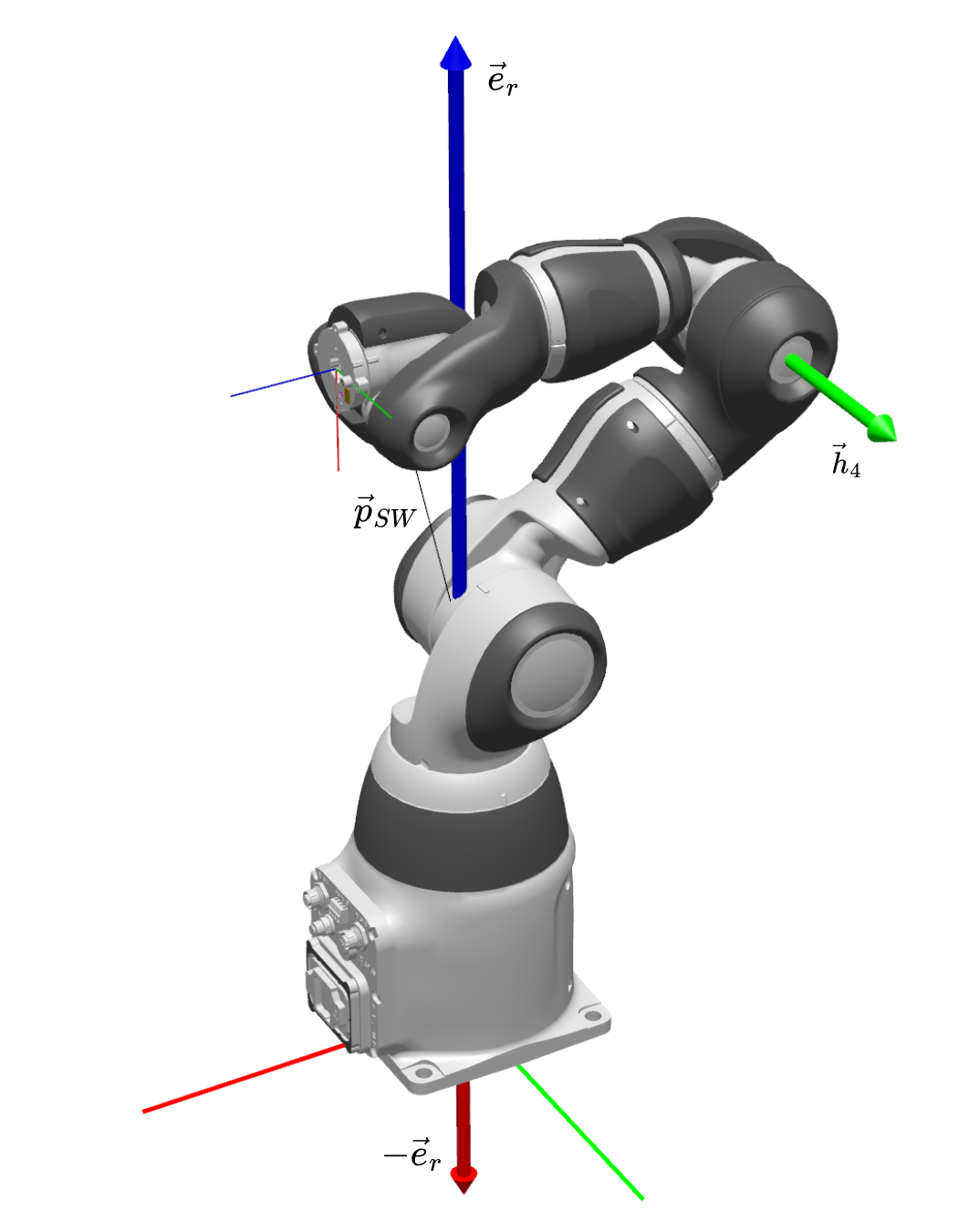}
    \caption{Singularity regions using the conventional and stereographic SEW angles. With the conventional SEW angle, the arm encounters a coordinate singularity when the wrist is on the line passing through the shoulder in the \(\vec e_r\) direction (blue and red arrows). With the stereographic SEW angle, the arm only encounters a coordinate singularity on a half-line. The half-line can be placed in the \(-\vec e_r\) direction (red arrow) which is inside the robot base and out of reach.}
    \label{fig:singularity_regions}
\end{figure}

Although the coordinate singularity is unavoidable for a 7-DOF robot such as the YuMi, the Stereographic SEW angle~\cite{elias2024_7dof} reduces this singularity from a full line in the workspace to a half-line which can be placed inside the robot base and out of reach. This is illustrated in Fig.~\ref{fig:singularity_regions}. Although the ABB controller does not support the Stereographic SEW angle parameterization directly, it could be implemented externally by streaming joint commands

Finally, a general redundancy parameterization may exhibit a fourth type of singularity where the parameterization Jacobian becomes zero. However, this does not occur for the SEW parameterization used by ABB.

\section{Inverse Kinematics}\label{sec:IK}

The IK problem is solved similarly to the example provided in~\cite[Sec.~6.2.10]{elias2024_7dof} which builds on the 6-DOF subproblem decomposition method shown in~\cite{elias2025_6dof}. The following is just one of the many ways to use subproblem decomposition to reduce the problem to a search over 2 angles. Since \(p_{01}\) is nonzero in the provided YuMi kinematic parameters, we will explicitly subtract \(p_{01}\) from \(p_{07}\).

For a given end effector pose \((R_{0T}, p_{0T})\), we can find \((R_{07}, p_{07})\). Then, given \(p_{SW} = p_{07}-p_{01}\) and conventional SEW angle~\(\psi^{conv}\), we find
\begin{equation}
    n_{SEW} = \rot(e_{SW},\psi^{conv}) e_{y,C} = \rot(e_{SW},\psi^{ABB})e_{x,C}
\end{equation}
where \(n_{SEW}\) is the normalized version of \(p_{SW}^\times R_{03}h_4\). This constraints \(R_{03}h_4\) to lie on a semicircle.

Given \((q_1, q_2)\), we use Subproblem~4 to find \(q_3\) such that \(R_{03} h_4\) is normal to \(n_{SEW}\):
\begin{equation}
    n_{SEW}\tr R_{02}R_{23}h_4 = 0
\end{equation}

We only keep the solutions which keep \(R_{03} h_4\) in the correct half-plane:
\begin{equation}
    e_{CE}\tr R_{03}h_4 > 0,
\end{equation}
where \(e_{CE}=n_{SEW}^\times e_{SW}\).

The remainder of the IK solution matches the example shown in~\cite{elias2024_7dof}. Find \((q_5, q_6, q_7)\) using Subproblem~5 to solve
\begin{equation}
-p_{67} + R_{67} R_{07}\tr (p_{07}-p_{01}-p_{14}) = R_{65} (p_{56} + R_{54} p_{45}),
\end{equation}
and find the error
\begin{equation}
    e(q_1,q_2) = \norm{R_{03}\tr R_{07} R_{47}\tr h_4 - h_4}.
\end{equation}
Search over \((q_1, q_2)\) to find zeros of this error. Then, find \(q_4\) using Subproblem~1:
\begin{equation}
    R_{34}p = R_{03}\tr R_{07} R_{47}\tr p,
\end{equation}
where \(p\) is any vector not collinear with \(h_4\).

To demonstrate the flexibility of the subproblem decomposition approach, we provide an alternative IK solution that searches over \(q_1\) and \(q_6\) and finds the error in SEW angle. For each fixed \(q_1\), the 7-DOF manipulator becomes a 6-DOF manipulator with non-consecutive intersecting joint axes 5 and 7. Using the solution method provided in~\cite{elias2025_6dof}, a 1D search can be performed over \(q_6\) to find all IK solutions with the correct end effector pose. This search creates a plot of \(\psi^{conv}\) versus \(q_1\) for a fixed end effector pose. The IK solutions are the intersections of this plot with the desired \(\psi^{conv}\). The resulting plot is also a useful graphical tool to determine how the IK solutions change depending on the chosen SEW angle.

\section{Testing and Comparison}\label{sec:testing}
\begin{table*}[t]
    \centering
    \caption{Comparison of SEW angles from RobotStudio and using both analytical expressions. All units are in degrees and shown to two decimal places to match the units and precision of RobotStudio. Discrepancies are shown in bold.}
\begin{tabular}{rrrrrrrrrrrrrr}
\toprule
   &    &    &    &    &    &    &    & \multicolumn{3}{c}{\(\vec e_r = \vec e_{z,0}\)} & \multicolumn{3}{c}{\(\vec e_r = \vec e_{y,0}\)} \\ \cmidrule(r){9-11} \cmidrule(l){12-14} 
\# & \(q_1\) & \(q_2\) &  \(q_3\) & \(q_4\) & \(q_5\) & \(q_6\) & \(q_7\) & \(\psi^{RS}\)   & \(\psi^{ABB}\)   & \(\psi^{sign}\)   & \(\psi^{RS}\)   & \(\psi^{ABB}\)   & \(\psi^{sign}\)   \\ \midrule
1  &     0.00 &     0.00 &     0.00 &     0.00 &     0.00 &     0.00 &     0.00 &     0.00 &     0.00 &     0.00 &    90.00 &    90.00 &    90.00 \\
2  &    20.00 &    20.00 &    20.00 &    20.00 &    20.00 &    20.00 &    20.00 &     6.86 &     6.86 &     6.86 &   105.54 &   105.54 &   105.54 \\
3  &    30.00 &    30.00 &    30.00 &    30.00 &    30.00 &    30.00 &    30.00 &    14.50 &    14.50 &    14.50 &    99.26 &    99.26 &    99.26 \\
4  &   -40.00 &   -40.00 &   -40.00 &   -40.00 &   -40.00 &   -40.00 &   -40.00 &   106.67 &   106.67 &   106.67 &    15.68 &    15.68 &    15.68 \\
5  &   -80.00 &   -80.00 &   -80.00 &   -80.00 &   -80.00 &   -80.00 &   -80.00 &    99.01 &    99.01 &    99.01 &   -26.00 &   -26.00 &   -26.00 \\
6  &   104.00 &   -31.00 &     1.00 &   -18.00 &   -55.00 &   117.00 &    69.00 &    19.58 &    19.58 & \textbf{  -19.92} &  -165.56 &  -165.56 & \textbf{ -165.08} \\
7  &   162.00 &     1.00 &   -23.00 &   -88.00 &   241.00 &    98.00 &   -44.00 &    80.07 &    80.07 & \textbf{  -80.08} &  -131.03 &  -131.03 & \textbf{ -130.97} \\
8  &   -20.00 &  -139.00 &  -144.00 &    77.00 &  -263.00 &   128.00 &  -104.00 &   139.68 &   139.68 & \textbf{  128.78} &   -15.92 &   -15.92 & \textbf{  -37.81} \\
9  &    67.00 &  -142.00 &   121.00 &    78.00 &   -38.00 &   113.00 &  -194.00 &  -146.69 &  -146.69 & \textbf{ -130.33} &   119.74 &   119.74 & \textbf{  112.59} \\
10 &   -58.00 &  -138.00 &   -35.00 &    70.00 &  -283.00 &   133.00 &   207.00 &   165.86 &   165.86 & \textbf{  156.08} &    50.38 &    50.38 & \textbf{   53.05} \\
\bottomrule
\end{tabular}
    
    \label{tab:SEW_comparison}
\end{table*}
\subsection{SEW Angle Definitions}
The SEW angle definition proposed in this work \(\psi^{ABB}\) was compared to the definition proposed in \cite{asgari2025singularities} \(\psi^{sign}\) and the SEW angle read directly from RobotStudio \(\psi^{RS}\). Results for 10 different joint configurations are shown in Table~\ref{tab:SEW_comparison}. SEW angles were computed using reference vectors pointing in the \(z\) and \(y\) directions. The first 5 test cases show a match among all three SEW angles. (Approximately 40\% of randomly chosen poses within the joint limits had matching SEW angles up to 0.01~deg away after rounding to 2 decimal places.) However, the next 5 hand-picked test cases show large discrepancies for \(\psi^{sign}\), while \(\psi^{ABB}\) continues to match \(\psi^{RS}\).

\subsection{Inverse Kinematics Solution}
\setlength{\tabcolsep}{4pt}
% \the\tabcolsep
\begin{table}[t]
    \centering
    \caption{IK solutions in degrees for a single pose with \(\vec e_r = \vec e_{z,0}\). Bold rows are within joint limits and are provided in RobotStudio.}
    \begin{tabular}{r *{7}{r}}
    \toprule
         \# & \(q_1\) & \(q_2\) &  \(q_3\) & \(q_4\) & \(q_5\) & \(q_6\) & \(q_7\) \\ \midrule
\textbf{ 1} & \textbf{ -16.60} & \textbf{  -8.10} & \textbf{ -93.20} & \textbf{  79.10} & \textbf{ 175.80} & \textbf{  37.80} & \textbf{ 199.80} \\
\textbf{ 2} & \textbf{  50.03} & \textbf{ -22.30} & \textbf{-158.19} & \textbf{  73.91} & \textbf{  -3.04} & \textbf{ -52.99} & \textbf{  18.42} \\
\textbf{ 3} & \textbf{-143.33} & \textbf{  14.30} & \textbf{  35.02} & \textbf{  74.48} & \textbf{  -2.89} & \textbf{ -44.59} & \textbf{  18.94} \\
\textbf{ 4} & \textbf{ 141.96} & \textbf{   8.55} & \textbf{ 108.30} & \textbf{  79.03} & \textbf{ 175.87} & \textbf{  34.61} & \textbf{-159.95} \\
 5  &  140.29 & -175.83 &  -78.44 &  128.02 &   27.80 &  100.51 &    1.51 \\
 6  &   48.97 &  165.70 &  138.57 &  118.44 &  -12.79 &   90.62 &   27.87 \\
 7  &  -69.39 &  160.40 &   28.21 &  140.63 &   -8.36 &   96.90 &   23.53 \\
 8  &   16.02 &  170.26 &   95.05 &  124.93 &  -21.65 &   94.07 &   33.72 \\
 9  &  -92.03 &  144.39 &   14.84 &  160.33 &  178.05 &  -95.36 & -160.37 \\
10  &   92.11 & -167.24 & -146.75 &  135.87 & -168.78 &  -98.68 & -168.22 \\

         \bottomrule
    \end{tabular}
    
    \label{tab:IK_solutions}
\end{table}
\setlength{\tabcolsep}{6pt}

Testing over random poses suggests there are up to 8 IK solutions within the joint limits and up to 14 IK solutions ignoring joint limits. Since the configuration numbering in RobotStudio goes from 0 to 7 and there are seemingly up to 8 IK solutions within the joint limits, there is no immediate indication that there are missed IK solutions in RobotStudio as is the case for the ABB GoFa~\cite{elias2025path}.

\begin{figure}[t]
    \centering
    \includegraphics[scale=0.5]{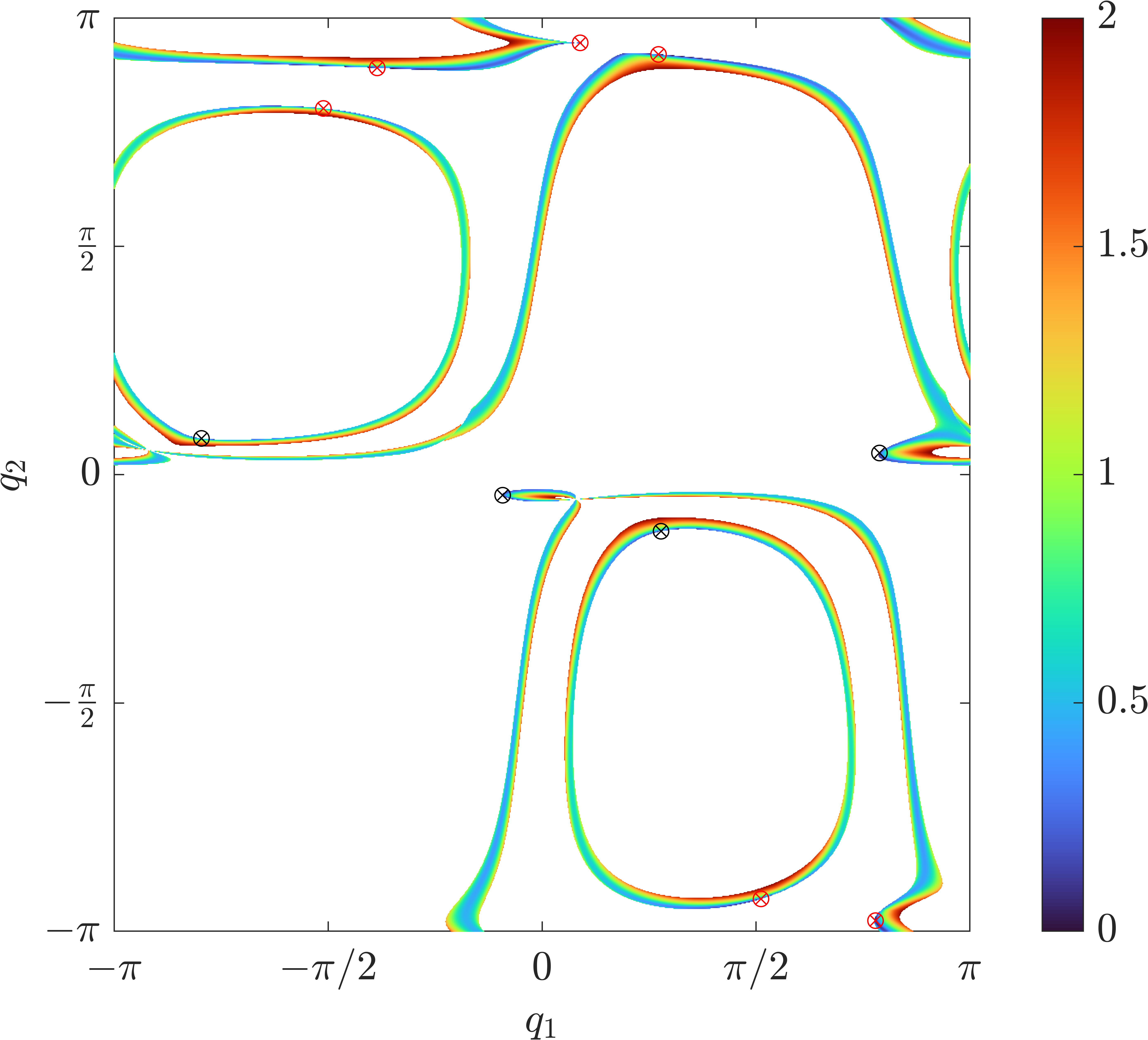}
    \caption{Finding all IK solutions using 2D search. The error function is multi-valued, and color indicates the minimum error among the four branches. The 10 IK solutions for this end effector pose and SEW angle occur where the error is zero and are marked with circled crosses. Black crosses lie within joint limits; red crosses lie outside.}
    \label{fig:plot_2D_search}
\end{figure}

\begin{figure}[t]
    \centering
    \includegraphics[scale=0.5]{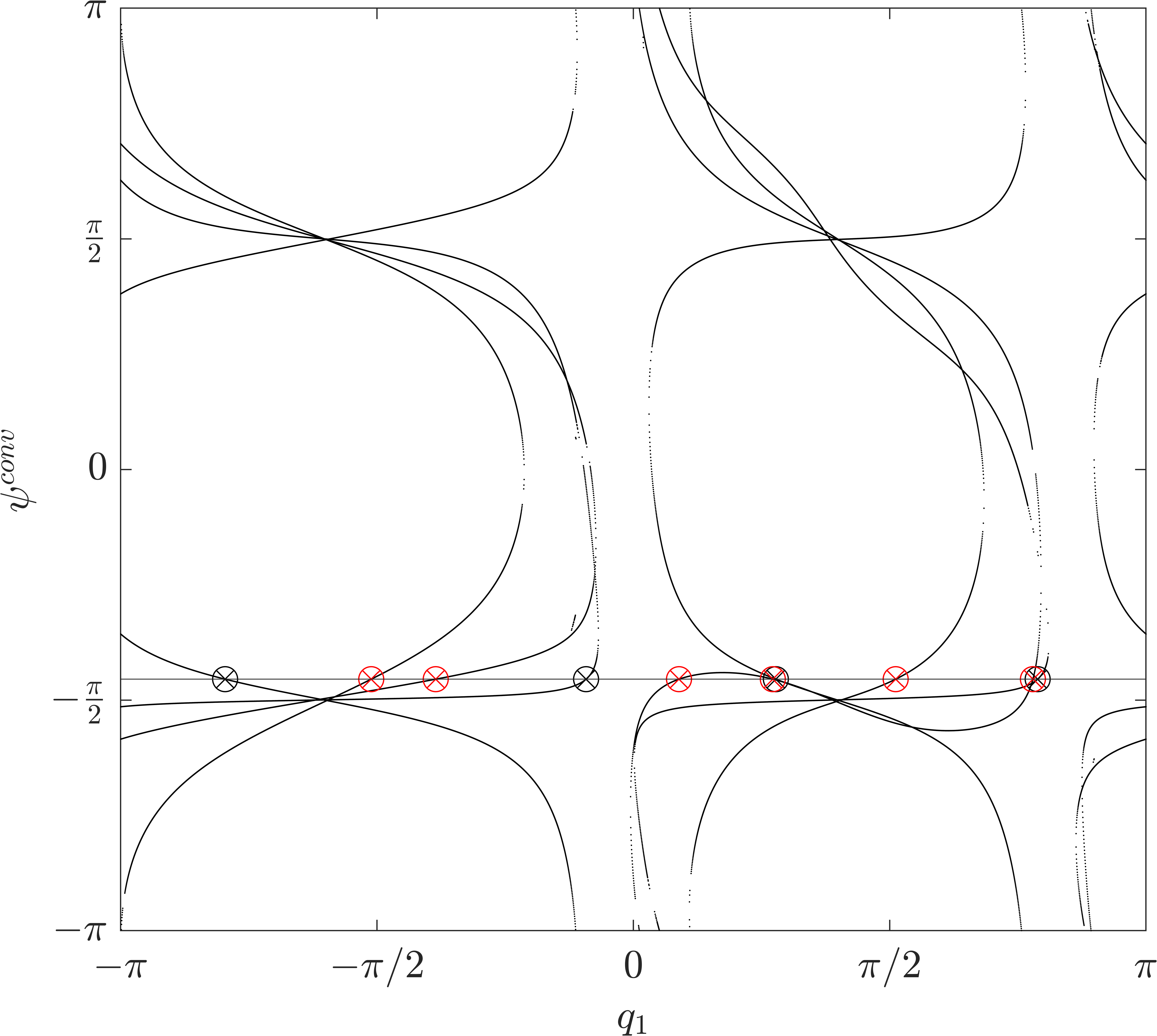}
    \caption{Finding all IK solutions using nested 1D search. For each \(q_1\), all IK solutions achieving the desired end-effector pose are found. The 10 IK solutions lie on the horizontal line representing the desired SEW angle are marked with circled crosses. Black crosses lie within joint limits; red crosses lie outside.}
    \label{fig:IK_nested_1D_search}
\end{figure}

All IK solutions for one end effector pose and SEW angle are shown in Table~\ref{tab:IK_solutions}. The corresponding 2D search plot is shown in Fig.~\ref{fig:plot_2D_search}. The alternative nested 1D search plot is shown in Fig.~\ref{fig:IK_nested_1D_search}. This example has 10 IK solutions, 4 of which are within the joint limits and match with the IK solutions shown in RobotStudio. The remaining 6 IK solutions fall outside the joint limits and were verified using forward kinematics. Although there are poses with up to 14 IK solutions, an example with 10 was chosen so that there would be a variety of solutions falling inside and outside the joint limits.


Finding all IK solutions in RobotStudio and IK-Geo take approximately the same amount of time provided the 2D search interval is set small enough to have a strong likelihood of finding all solutions. Increasing the search interval means the execution time is faster but missed solutions are more likely.

RobotStudio sometimes returns solutions which have the correct end effector pose but the incorrect SEW angle. This is easily checked, but can be confusing to inexperienced users. Additionally, the ``Limit avoidance distance" controller parameter appears to affect the solution accuracy in RobotStudio.

\section{Conclusion}\label{sec:conclusion}

The ABB YuMi is a particularly challenging arm for path planning, and a critical component of any planner is the ability to perform forward and inverse kinematics. Until now, it was not possible to exactly replicate the ABB controller’s kinematics because the definition of the SEW angle was not known. By identifying the SEW angle definition and verifying its correctness, we are able to perform forward kinematics through closed-form computation and inverse kinematics through 2D search.

With this complete kinematic description, we take an important step toward fully understanding the singularity structure of the YuMi. Although all kinematic singularities were identified in~\cite{asgari2025singularities}, the SEW angle definition provides clarity on the algorithmic singularities. These include SEW angle singularities, where the SEW angle is undefined, and augmentation singularities, where self-motion does not change the SEW angle.

We propose several next steps to continue the kinematic analysis of the YuMi arm.

One important task is to comprehensively identify all algorithmic singularities. Currently, we can identify algorithmic singularities and numerically search for poses with augmentation singularities, but a systematic classification is still needed. Ideally, such a classification would carry geometric meaning, as was done for kinematic singularities~ in~\cite{asgari2025singularities}.

Another important direction is to demonstrate solving IK using the polynomial method, where solutions correspond to the roots of a high-degree polynomial in the tangent half-angle of one joint. The authors of \cite{asgari2025singularities} mentioned there is a need for an analytical IK solver for the YuMi.
The current search-based IK procedure can likely be extended to use the polynomial method, as was demonstrated in \cite{elias2025_6dof, elias2024_7dof}. Although 2D search works well with a sufficiently small sampling interval, converting the problem to polynomial root finding provides stronger guarantees of finding all solutions.

Modern robot arms are increasingly moving away from the standard 6-DOF spherical wrist design. Instead, 7-DOF robots with fewer intersecting or parallel joints are becoming more common.
Path planning and the design of the underlying robot controller, including the redundancy parameterization as a deliberate design choice, require a complete understanding of the resulting algorithmic singularities.
We therefore hope this work in analyzing the YuMi SEW angle provides guidance not only for path planning with the YuMi but also inspires future work in understanding redundancy and influencing the design of next-generation arms and their controllers.

\section*{Acknowledgment}
The authors appreciate the generous and engaging discussion with Prof. I. A. Bonev and thank him for his time and interest in the results of this work.
\bibliographystyle{ieeetr}
\bibliography{bib}
\end{document}